\documentclass[conference]{IEEEtran}
\IEEEoverridecommandlockouts

\usepackage{cite}
\usepackage{amsmath,amssymb,amsfonts}
\usepackage[font=small,labelfont=bf]{caption}

\usepackage{subcaption}
\usepackage{graphicx}
\usepackage{textcomp}
\usepackage{xcolor}
\usepackage{tabularx}
\usepackage{balance}
\usepackage{hyperref}
\usepackage{tikz}
\usetikzlibrary{positioning}
\usepackage{listings}
\usepackage{booktabs}

\usepackage[left=19.1mm,right=19.1mm,top=25.4mm,bottom=19.1mm]{geometry}

\usepackage[ruled,linesnumbered]{algorithm2e}
\RestyleAlgo{ruled}
\SetKwComment{Comment}{/* }{ */}

\def\X{X}

\def\H{H}

\def\B{B}

\def\oldmethod{Original-PTO\xspace}

\def\method{PTO\xspace}

\newcommand{\sqboxs}{1.5ex}
\newcommand{\sqbox}[1]{\textcolor{#1}{\rule{\sqboxs}{\sqboxs}}}

\def\Gbelief{\ensuremath{G_{\text{belief}}}}
\def\Grandom{\ensuremath{G_{\text{random}}}}
\def\pathtree{\psi}

\title{Asymptotically Optimal Belief Space Planning in Discrete Partially-Observable Domains}
\title{PTO*: Asymptotically Optimal Belief Space Planning in Discrete Partially-Observable Domains}
\title{Asymptotically Optimal Belief Space Planning in Discrete Partially-Observable Domains}

\author{Janis Eric Freund$^{1}$, Camille Phiquepal$^{2}$, Andreas Orthey$^{1, 3}$, Marc Toussaint$^{1}$%
\thanks{$^{1}$Technical University of Berlin, Germany}%
\thanks{$^{2}$Machine Learning \& Robotics Lab, University of Stuttgart, Germany}%
\thanks{$^{3}$Realtime Robotics Inc., Boston, MA, USA}%
}

\begin{document}
\maketitle

\pagenumbering{arabic}
\thispagestyle{plain}
\pagestyle{plain}

\begin{abstract}

Robots often have to operate in discrete partially observable worlds, where the states of world are only observable at runtime. 
To react to different world states, robots need contingencies.
However, computing contingencies is costly and often non-optimal.
To address this problem, we develop the improved path tree optimization (PTO) method.
PTO computes motion contingencies by constructing a tree of motion paths in belief space. 
This is achieved by constructing a graph of configurations, then adding observation edges to extend the graph to belief space.
Afterwards, we use a dynamic programming step to extract the path tree. 
PTO extends prior work by adding a camera-based state sampler to improve the search for observation points. 
We also add support to non-euclidean state spaces, provide an implementation in the open motion planning library (OMPL), and evaluate PTO on four realistic scenarios with a virtual camera in up to 10-dimensional state spaces. 
We compare PTO with a default and with the new camera-based state sampler. The results indicate that the camera-based state sampler improves success rates in 3 out of 4 scenarios while having a significant lower memory footprint.
This makes PTO an important contribution to advance the state-of-the-art for discrete belief space planning.

\end{abstract}

\section{Introduction}

Motion planning is a prerequisite to use robots in search and rescue missions, and for autonomous driving. However, in real-world environments, a robot might have limited sensing ability and can only observe the world partially. While robots might have a precise map of an office building, they might not know if doors are open or closed. This requires robots to make observations to figure out the state of the world. 
However, once a robot makes an observation, it needs to potentially replan its path to fulfill its task. This is often costly and sub-optimal. 
\begin{figure}
    \centering
    \includegraphics[width=0.95\linewidth]{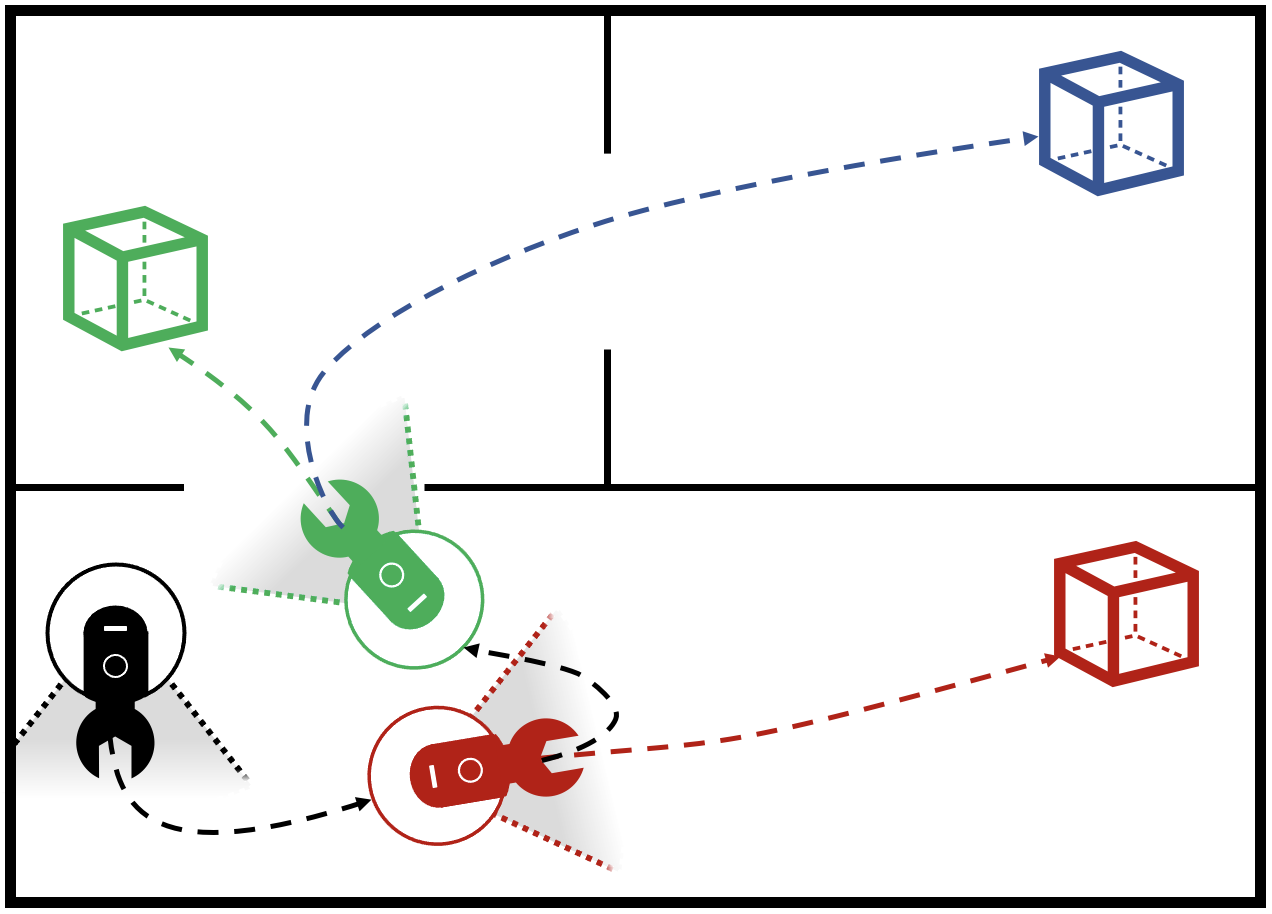}
    \caption{A planning problem, where the robot (black) needs to find a partially observable box which can be located at three different locations (green, red, blue). The robot can only observe the locations when they are in its field of view (grey). Our planner \method computes contingencies for robot motions, such that the robot can act optimally, even when the robot either observes the object at a location (green, red robot poses), or if the object is not present.}
    \label{fig:enter-label}
\end{figure}

In prior work~\cite{pathtree}, this problem was addressed by computing a path tree, which captures contingencies for how to act when observing different world states. The resulting path tree optimization method~\cite{pathtree} (\oldmethod) showed promising results in realistic scenarios. 
However, \oldmethod uses a simple observation model, and no observations from a realistic camera image. 
\oldmethod also does not support non-euclidean configuration spaces, which makes it not applicable to a wide set of robots. 
\oldmethod is also not implemented in the open motion planning library (OMPL), which makes it difficult to compare to other planners and to disseminate the method to the wider research community. Furthermore, samples in \oldmethod are generated randomly, which makes it hard to find the crucial observation points. 

To address those problems, we develop an extended and improved version of the path tree optimization (\method) planner. \method works similar to \oldmethod by planning a path tree in belief space, where a path branches at observation points that correspond to information gains of the environment. However, \method differs by being fully implemented in OMPL, supporting both non-euclidean configuration spaces, and a simulated camera model. Furthermore, a new camera-based state sampler is developed, which generates higher quality samples by sampling configurations where the camera points into the general direction of partially-observable objects. In summary, our contributions are:

\begin{enumerate}
    \item An implementation of \method as part of the Open Motion Planning Library (OMPL)\cite{ompl}, including a virtual camera model, and an extension to non-euclidean state spaces.
    \item A new camera-based state sampler, which samples states that lead to a higher chance of making belief-changing observations.
    \item An evaluation on four scenarios by integrating the OMPL implementation into the PyBullet~\cite{pybullet} simulator.
\end{enumerate}
\begin{figure*}[t]
    \centering
    \includegraphics[width=0.9\linewidth]{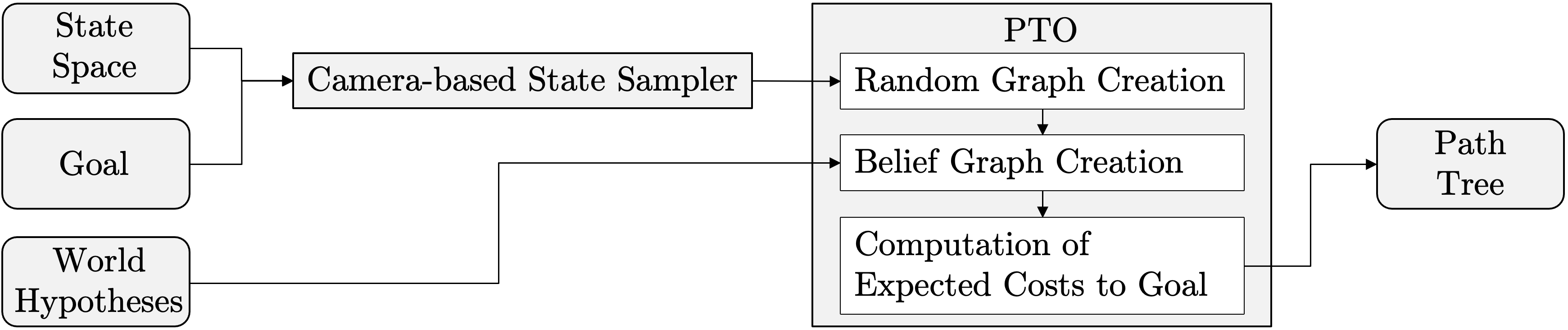}
    \caption{Structure of the PTO algorithm. The respective algorithm parts of PTO are explained in sections \ref{sec:create_random}, \ref{sec:create_belief_graph}, and \ref{sec:pathtree}. 
    \label{fig:system}}
    \vspace*{-0.4cm}
\end{figure*}

\section{Related Work\label{sec:related}}
\vspace*{-0.5cm}

Most work on motion planning in belief space builds on classic sampling-based planners like the rapidly exploring random trees (RRTs)~\cite{rrt}, and probabilistic roadmaps (PRMs)~\cite{prm}. RRTs grow a random tree by sampling the configuration space iteratively. PRMs instead construct a roadmap graph of the free configuration space. 
While the base variants of RRTs and PRMs are not asymptotically optimal~\cite{rrtstar}, other variants like rapidly exploring random graphs (RRGs)~\cite{rrtstar}, optimal RRTs (RRT$^*$)~\cite{rrtstar}, and batch informed trees (BIT$^*$)~\cite{bitstar} are asymptotically optimal. 
Our approach differs to sampling-based planners by building upon RRGs and using them as the foundation to conduct planning in discrete belief space.

Belief spaces are often introduced because of uncertainty about the world. Motion planning under uncertainty can be formalized using partially observable Markov decision processes (POMDPs)~\cite{pomdp}. POMDPs extend Markov decision processes (MDPs) by including situations in which the robot does not know about the state of the world. There are algorithms that try to solve these problems using Gaussian belief spaces~\cite{brm, covariance, rrbt, gaussian_trees}. The goal of these algorithms is to minimize the uncertainty of the robot's location. 
To solve POMDPs, roadmaps in belief space can be planned to deal with uncertainties of motions and observations~\cite{brm, covariance}. 
Roadmap-based planning can be combined with dynamic replanning to react to changing environments and deviations from the position of the robot~\cite{mobile}. 
POMDPs have also been combined with a model predictive control step to account for uncertainties in the world state~\cite{driving}. Another area where POMDPs have been successfully applied are task and motion planning (TAMP) problems, where POMDPs have been used to reduce uncertainties~\cite{modular, replanning}, and to create paths across different modes corresponding to specific tasks~\cite{tamp}. 
Those approaches are complementary to our approach in that they plan in continuous belief spaces. 
Instead, we consider problems involving discrete beliefs about the state of the environment and finite observations.

The closest work is the \oldmethod method by Phiquepal et al.~\cite{pathtree}. Based upon earlier works~\cite{Phiquepal2021}, this planner first introduces the idea of planning path trees in belief space. Our \method method builds directly upon this work and adds substantial improvements as stated in our contributions. 
\section{Discrete Belief Space Planning\label{sec:methods}}
\label{sec:pogoals}

\begin{figure}
    \centering
    \includegraphics[width=0.48\columnwidth]{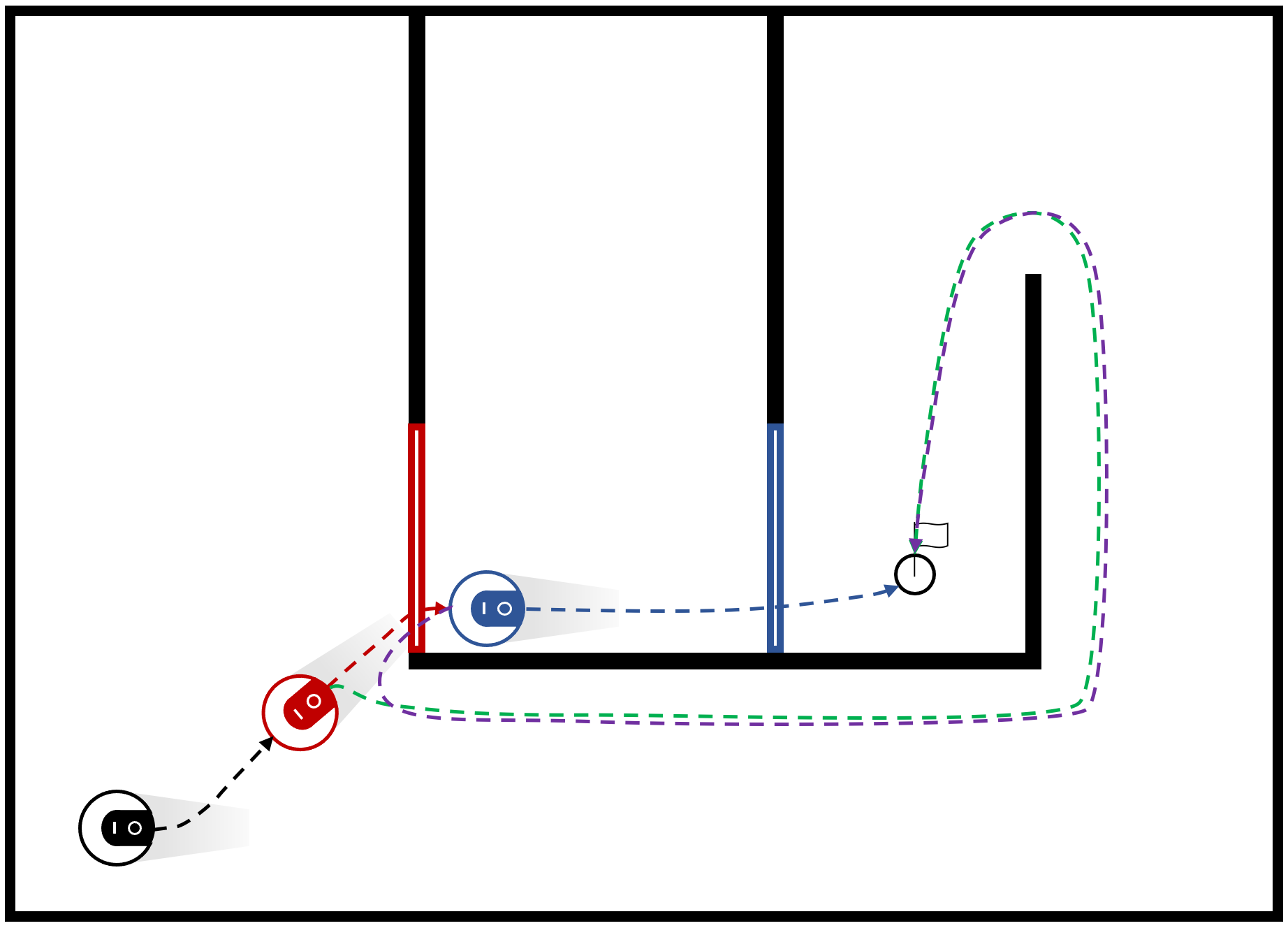}
    \includegraphics[width=0.48\columnwidth]{bilder/simple_goals.png}
    \caption{We support two world states. \textbf{Left:} A world with partially observable objects, namely two doors (red, blue). The robot starts at the black state and needs to reach the white flag. The resulting path tree has two observation points (red, blue), at which the robot observes the state of the doors (open or closed) and plans accordingly. \textbf{Right:} A world with a partially observable goal state (red, green, or blue box is present). The resulting path tree has two observation points (green, red), at which the robot updates its path according to the observation made (box exists or not).\label{fig:world_states}}
\end{figure}

A belief space planning problem for a robot $R$ is defined by a robot state space $\X=(Q,\B)$, consisting of the configuration space $Q$, and a belief space $\B$, together with a world $W$, and a hypothesis space $\H$. The robot configuration space $C$ is a fully-observable $n$-dimensional space representing robot configurations $q \in Q$. The world $W$ is an environment in which objects and goals can have different states, as depicted in Fig.~\ref{fig:world_states}. It is initially unknown in which state the world $W$ happens to be. 
The hypothesis space $\H$ is the finite, discrete set $h \in \H$ representing all possible states in which the world $W$ can be. 
Finally, the belief space $\B$ represents probability distributions $b \in \B$ over the world hypothesis space $\H$ with $\B$ being the set of all probability distributions on $\H$. 

The robot $R$ can explore the world through observations $o \in O$. An observation at state $x \in \X$ is a function from the camera image and the belief of the robot to a discrete output if an object state has been detected. We assume that observation outputs are binary, i.e. we either detect an object state or we do not. 

The goal of discrete belief space planning is to create a path tree $\pathtree$. A path tree $\pathtree$ is a directed tree, where edges are paths in the state space $\X$ having either the same belief state (movement edges), or the same configuration (observation edges). Nodes are observation points at which the belief state of the robot changes. This path tree represents contingencies for the robot, such that for every observation outcome (e.g. an object is present or not), the robot has a possible option of how to move in the world $W$. 

\subsection{Optimization Objective}

We are not only interested in finding some path tree, but in finding an optimal path tree $\pathtree^*$ over all possible path trees~\cite{pathtree}. Our optimization objective is constructed as follows:

\begin{subequations}
\begin{align}
& \pathtree^*=\underset{\pathtree}{\operatorname{argmin}} \sum_{(u, v) \in \pathtree} C(u, v) p\left(v \mid \pathtree, b_0\right), \label{eq:objective}\\
& \text { s.t. } \nonumber\\
& \forall h \in H, \exists l \in \mathcal{L}(\pathtree) \mid G(l), \label{eq:complete}\\
& \mathcal{V}(u, v), \forall(u, v) \in \pathtree, \label{eq:valid}\\
& b_v(h)=\frac{p(o \mid h) b_u(h)}{\sum_{h^{\prime}} p\left(o \mid h^{\prime}\right) b_u\left(h^{\prime}\right)}, \forall h \in H, (u, o, v) \in \pathtree\label{eq:update}
\end{align}
\end{subequations}

whereby $C(u, v)$ is the cost of the edge that connects the nodes $u$ and $v$. The objective defined by Eq.~\eqref{eq:objective} is to find an optimal path tree $\pathtree^*$ that minimizes the sum of the costs $C(u, v)$ over all pairs of nodes in a path tree $\pathtree$, while also taking into account the conditional probability of reaching the node $v$ given an initial belief $b_0$. 
Also, some constraints apply: Constraint \eqref{eq:complete} makes sure that for each state $h \in H$, it must exist at least one leaf node $l\in\mathcal{L}(\pathtree)$ that satisfies the goal condition given by $G(l)$. $\mathcal{L}(\pathtree)$ gives the set of all leaf nodes of the path tree $\pathtree$. This condition ensures the completeness of the path tree. Furthermore, the constraint \eqref{eq:valid} checks if all edges $(u, v) \in \pathtree$ are valid as given by $\mathcal{V}(u, v)$. The constraint \eqref{eq:update} specifies that the belief state $b_v(h)$ of the node $v$ should be consistent with the belief $b_u(h)$ of its parent node $u$ when making the observation $o$ while transitioning from $u$ to $v$. $p(o|h)$ is the observation model.

\section{Path Tree Optimization Method\label{sec:algorithm}}

The \method planner is composed of three steps as shown in Fig.~\ref{fig:system}. First, a random graph $\Grandom$ on the configuration space is created. This graph contains information about validity of nodes and edges in different world states. Second, a belief graph $\Gbelief$ on the state space (configuration space plus belief space) is created which adds edges on nodes where a belief change occurs. Finally, we use a dynamic programming search on the belief graph to compute optimal expected cost-to-go values for all nodes. Using those cost-to-go values, we eventually extract the path tree. 

\def\pWidth{0.32\linewidth}
\begin{figure*}
    \centering
    \begin{subfigure}{\pWidth}
        \includegraphics[width=\linewidth]{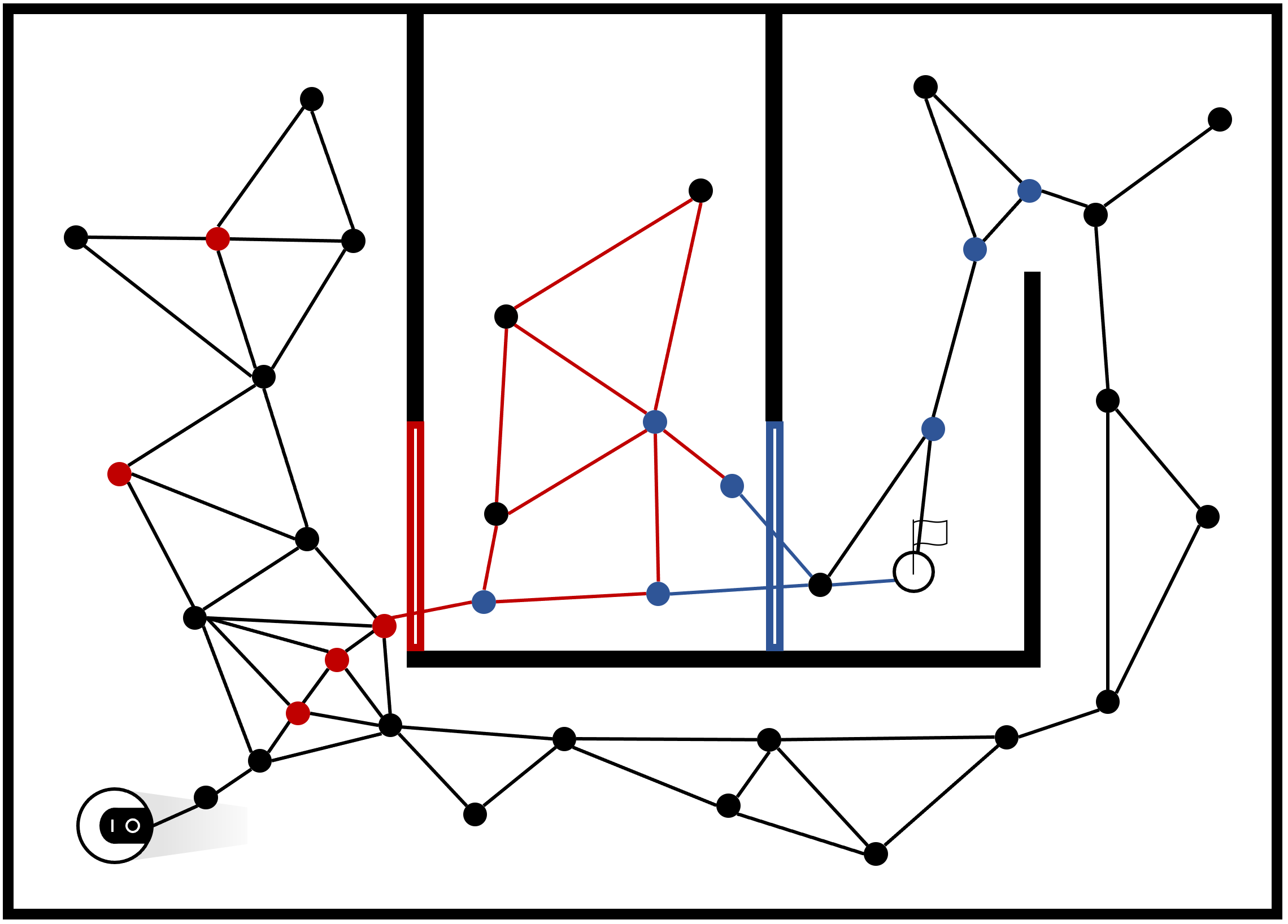}
        \caption{Creating a random graph $\Grandom$\label{fig:pto_random_graph}} 
    \end{subfigure}
    \begin{subfigure}{\pWidth}
        \includegraphics[width=\linewidth]{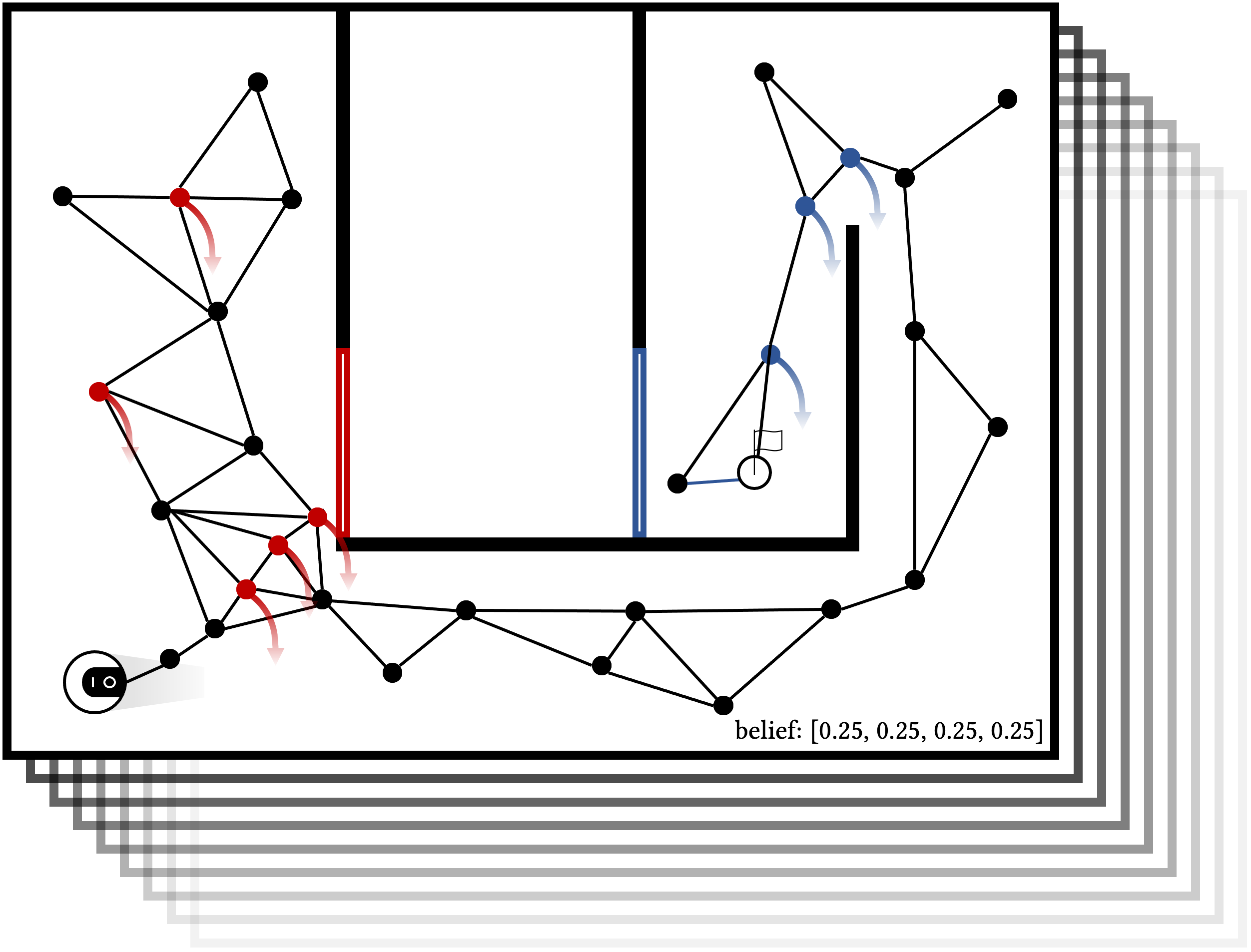}
        \caption{Creating a belief graph $\Gbelief$\label{fig:pto_belief_graph}} 
    \end{subfigure}
    \begin{subfigure}{\pWidth}
        \includegraphics[width=\linewidth]{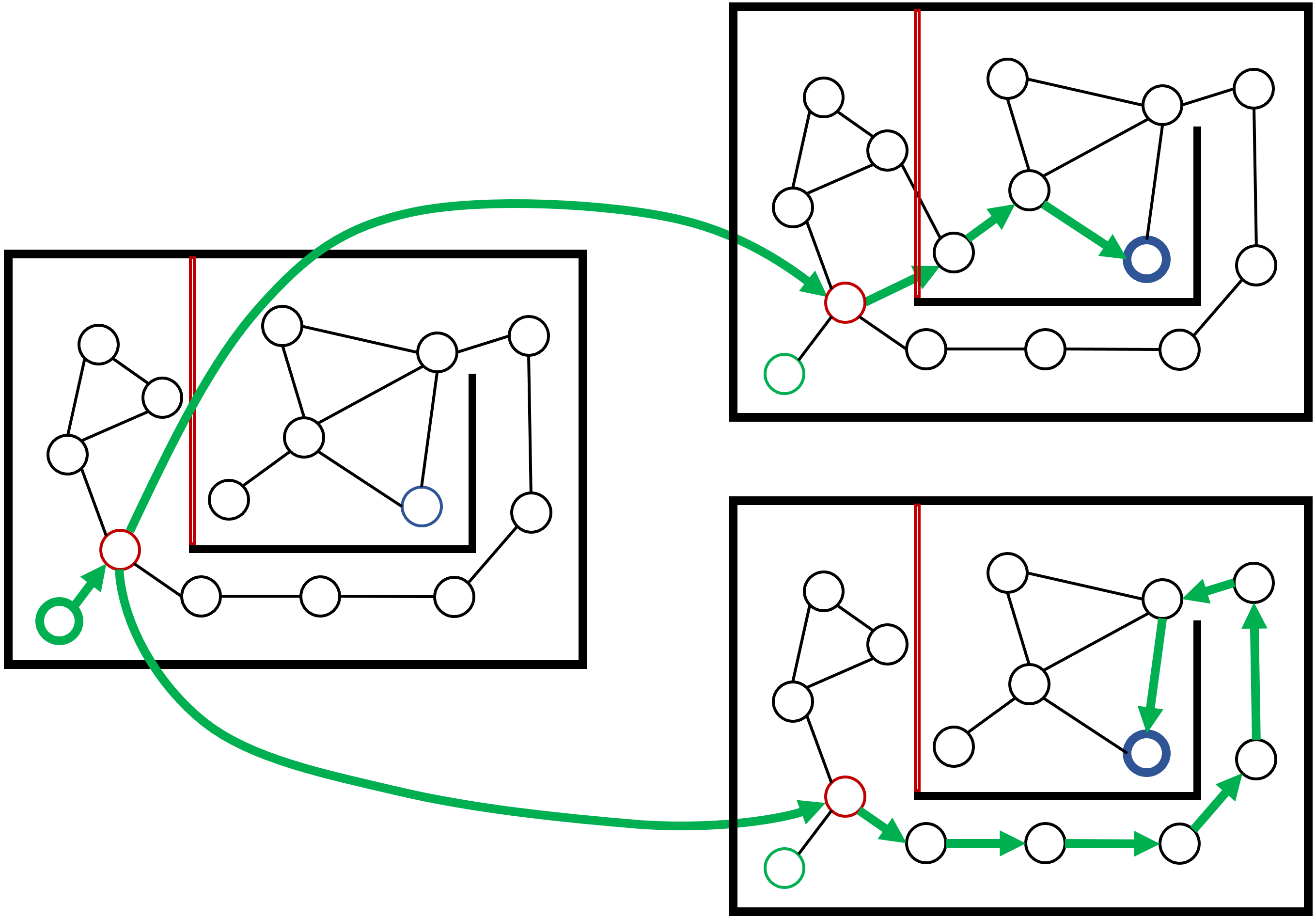}
        \caption{Extracting the path tree $\pathtree$\label{fig:pto_path_tree_extraction}} 
    \end{subfigure}
    \caption{The three steps of the PTO method. \textbf{Left:} A random graph is constructed on the configuration space, with information in which world states (~\sqbox{black} [1,2,3,4], ~\sqbox{red} [1,3], ~\sqbox{blue} [1]) an edge is valid. \textbf{Middle:} The random graph is extended to the belief space by adding edges whenever an observation changes the belief. \textbf{Right} Expected costs are calculated and the path tree is extracted from the belief graph.\label{fig:pto}}
\end{figure*}

\subsection{Random Graph Creation\label{sec:create_random}}

In the first step, a rapidly exploring random graph~\cite{rrtstar} $\Grandom$ on the robot configuration space is created. This graph is annotated with additional information: For each configuration, we store a list of all partially observable objects which can be seen. For each edge, we store a list of world states in which this edge is valid. 

The process of the random graph creation is illustrated in Fig.~\ref{fig:pto_random_graph}. 
It consists of two phases. First, we sample a random configuration in the robot configuration space, and we sample a random world state. It is then checked if the configuration is valid in this particular world state. 
If it is valid, we add it to the graph. Afterwards, we search for all observable objects in the world from the current configuration. This information is stored for use in the belief graph creation (Sec.~\ref{sec:create_belief_graph}).

Second, a neighborhood search is conducted to add edge connections to all viable neighbor configurations in the graph. This is accomplished by iterating over all world states, and getting the nearest configurations for each world state in a specific radius $R$. For each nearest configuration, we check if the edge is valid in the world state. If it is valid, we add the edge to the graph, or update an existing edge by adding the valid world state to the edge itself. 

This iterative process is stopped once a planner terminate condition is reached. This can be a number of iterations, or a timeout. The final random graph consists of nodes and edges on the configuration space, together with the information about valid world states. 

\subsection{Belief Graph Creation
\label{sec:create_belief_graph}}

In the second step, we use the random graph $\Grandom$ to create a belief space graph $\Gbelief$. This is illustrated in Fig.~\ref{fig:pto_belief_graph}. The belief graph $\Gbelief$ extends the random graph to the state space by creating nodes and edges, which both contain configurations and beliefs. 

The process of creating a belief graph consists of three phases. First, we iterate over all vertices $v$ in $\Grandom$, and over all belief states $b \in \B$. For each vertex belief pair $(v, b)$ we check if the vertex is reachable in the given belief $b$. This is accomplished by checking if there is at least one edge in which $b$ is valid. If there is at least one, we add the pair $(v, b)$ to the belief graph. 

In the second phase, nodes of the belief graph are connected with edges. Those edges are created by iterating over all edges on the random graph $\Grandom$. For each edge, we compute the compatible beliefs for the valid world states. For each compatible belief, we add one edge to the belief graph, which connects the nodes of the same belief. After the second phase, there is one graph for each belief, and those graphs are disconnected. 

In the third phase, all individual belief graphs are connected to each other. This means we add observation edges, where only the belief, but not the configuration changes. For this, we iterate over all vertices in the current belief graph $\Gbelief$. For each vertex, we analyse the observable objects from this configuration, and we compute the beliefs that result if the objects are observed at the given belief. For each of those beliefs, we then add an observation edge. For example, if there is an observable object at the configuration, we would add one connection to the belief state where the door is observed as closed, and one edge to the belief state where the door is observed to be open.

\subsection{Extraction of Path Tree\label{sec:pathtree}}

Once the belief graph $\Gbelief$ is constructed, we use a Dijkstra-like algorithm~\cite{dijkstra} to compute the optimal cost-to-go values for each vertex, from which we extract the path tree $\pathtree$. The cost-to-go is computed using dynamic programming by Bellmann updates for each vertex in the graph to all goal vertices. 

The path tree extraction is done recursively. Given the belief graph $\Gbelief$ with cost-to-go values, we first compute a list containing all goal vertices. Those vertices are added to the path tree. Then, for each goal vertex, we add the next best node with the lowest cost-to-go value. This process is continued until we either reach the start vertex, or we reach a branching point, where two paths are joined along a single vertex. Once the method terminates, we return the path tree which can then be used by the robot to move through the world towards the goal.
\def\figWidth{\linewidth}
\def\figHeight{0.62\linewidth}
\def\subfigWidth{0.32\linewidth}
\begin{figure*}[t]
    \centering
    \begin{subfigure}{\subfigWidth}    
        \includegraphics[width=\figWidth, height=\figHeight]{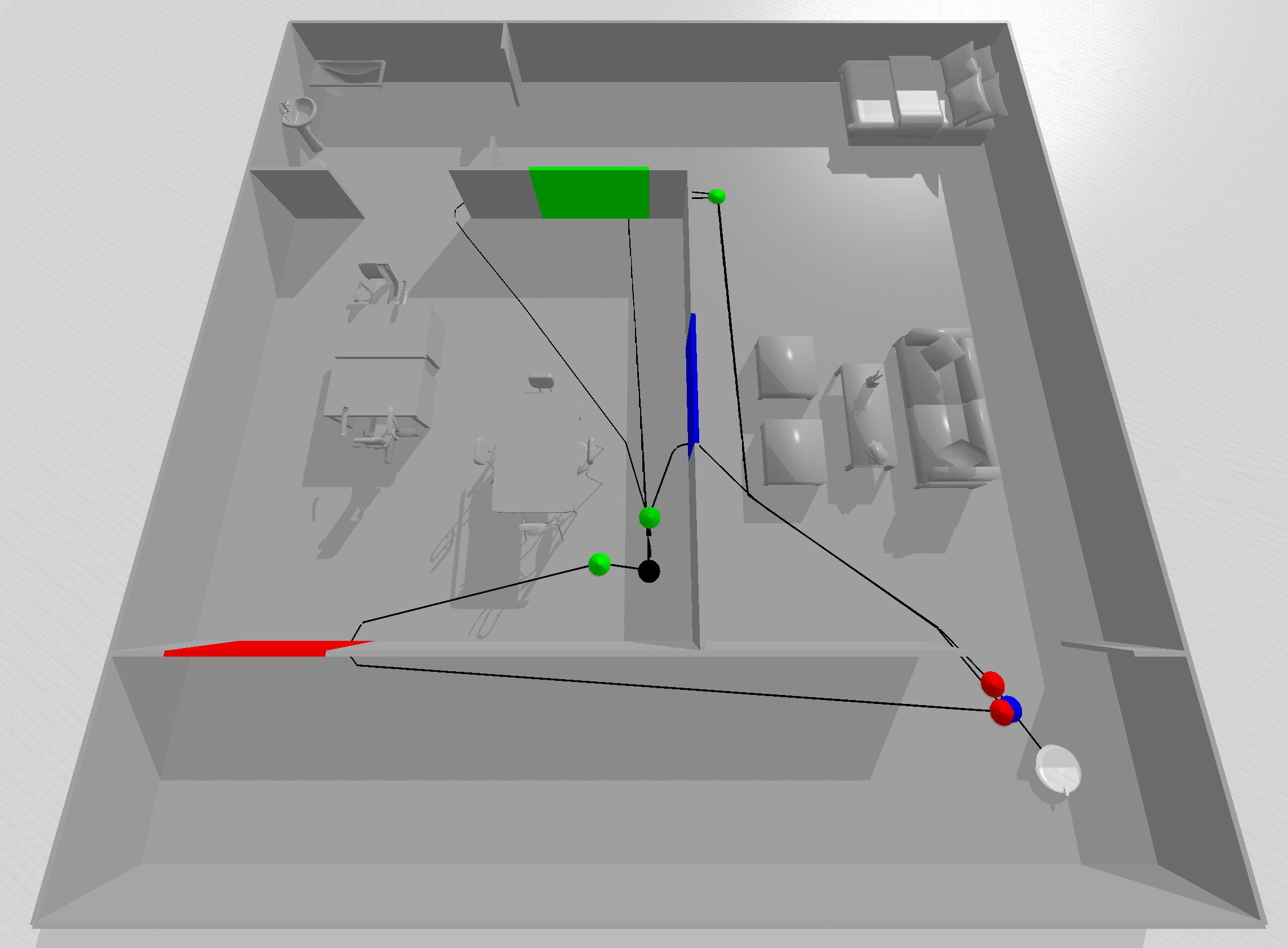}
        \caption{Door Environment\label{fig:door_environment}}        
    \end{subfigure}
    \hfill
    \begin{subfigure}{\subfigWidth}    
    \includegraphics[width=\figWidth, height=\figHeight]{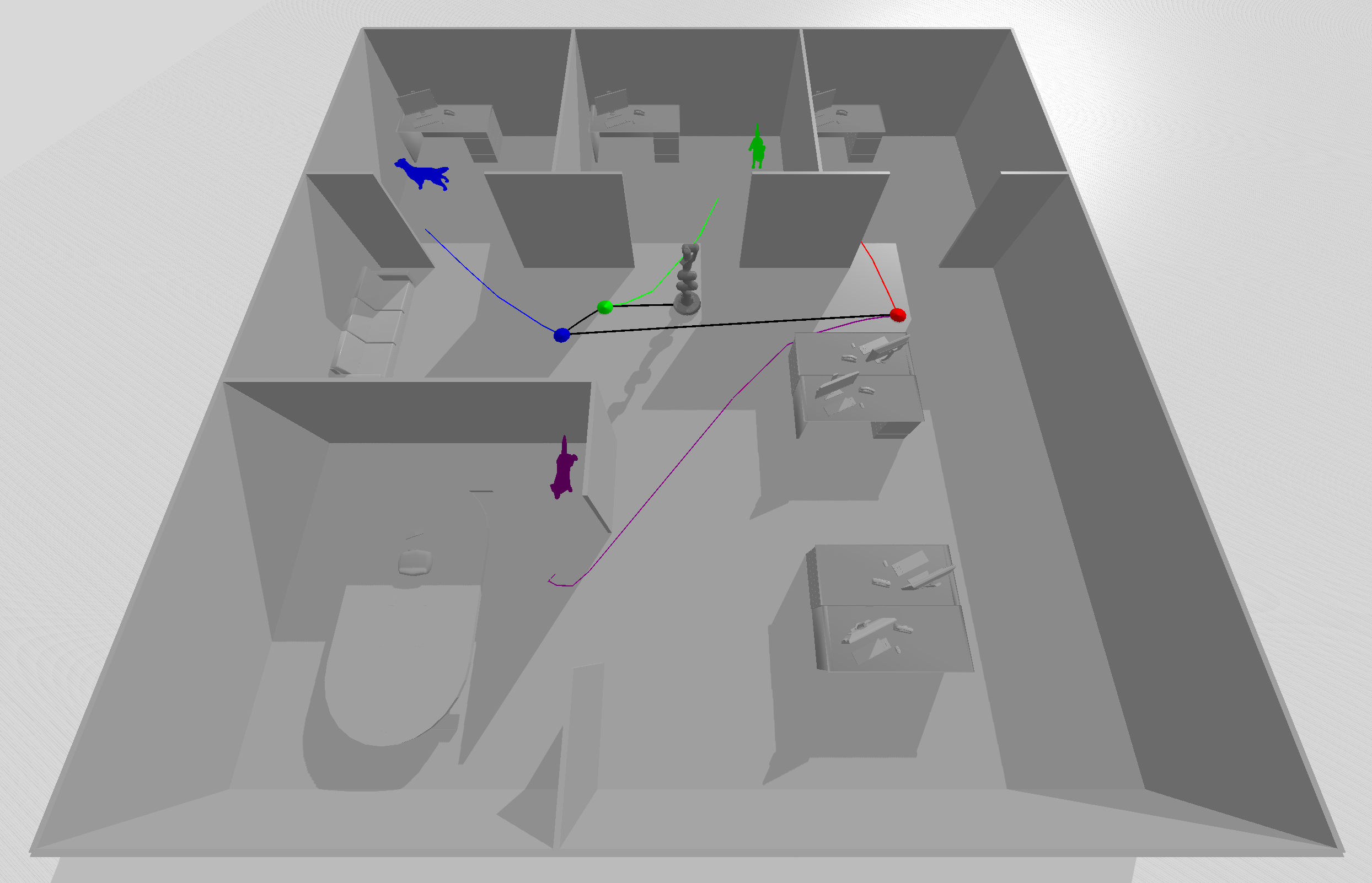}
        \caption{Search and Rescue Environment\label{fig:sar_environment}}        
    \end{subfigure}
    \hfill
    \begin{subfigure}{\subfigWidth}    
        \includegraphics[width=\figWidth, height=\figHeight]{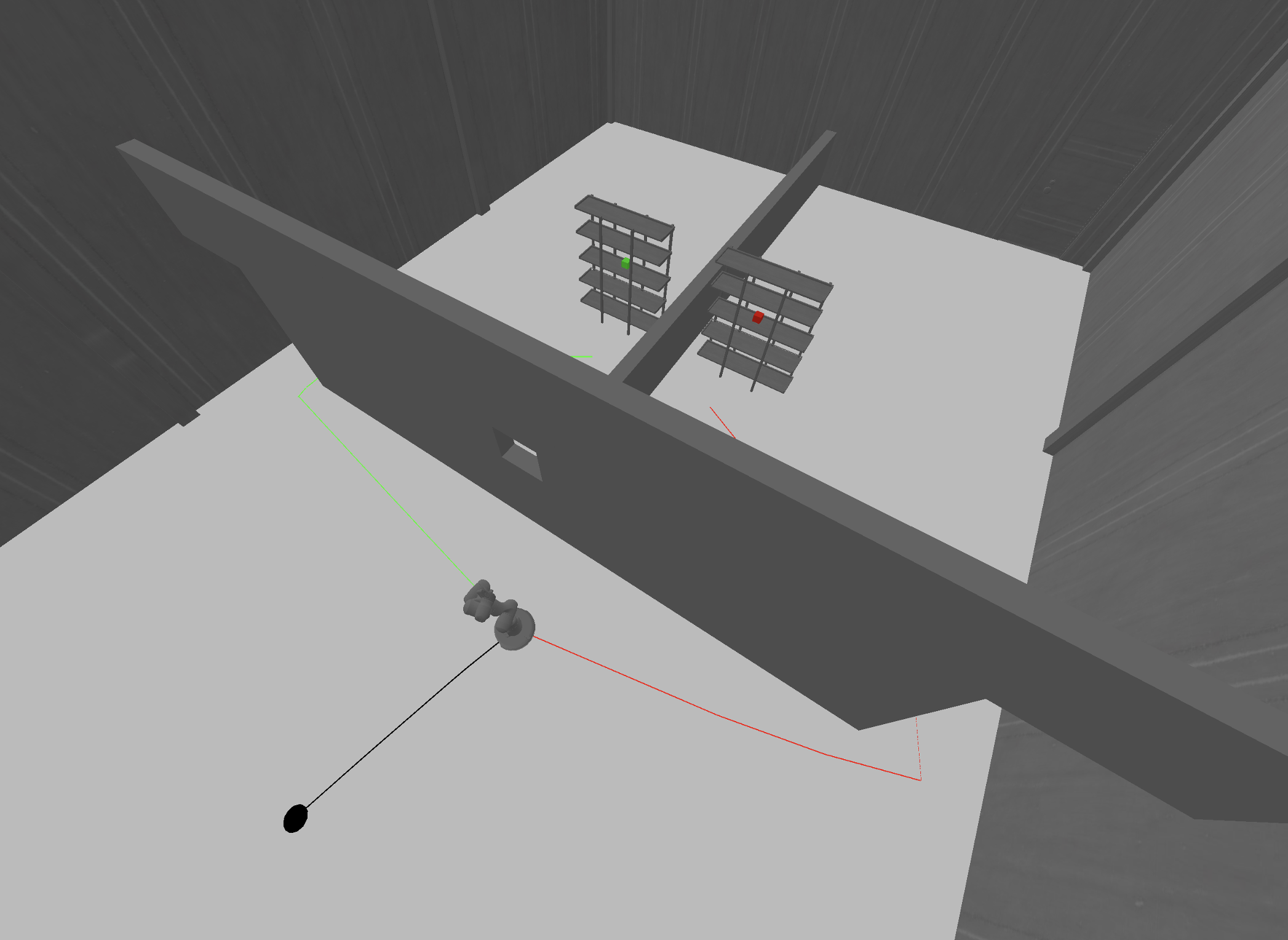}
        \caption{Warehouse Environment\label{fig:warehouse_environment}}        
    \end{subfigure}
    \caption{The door, search and rescue, and warehouse environments used for benchmarking.}
\end{figure*}

\begin{figure}
    \centering
    \includegraphics[width=1.0\linewidth]{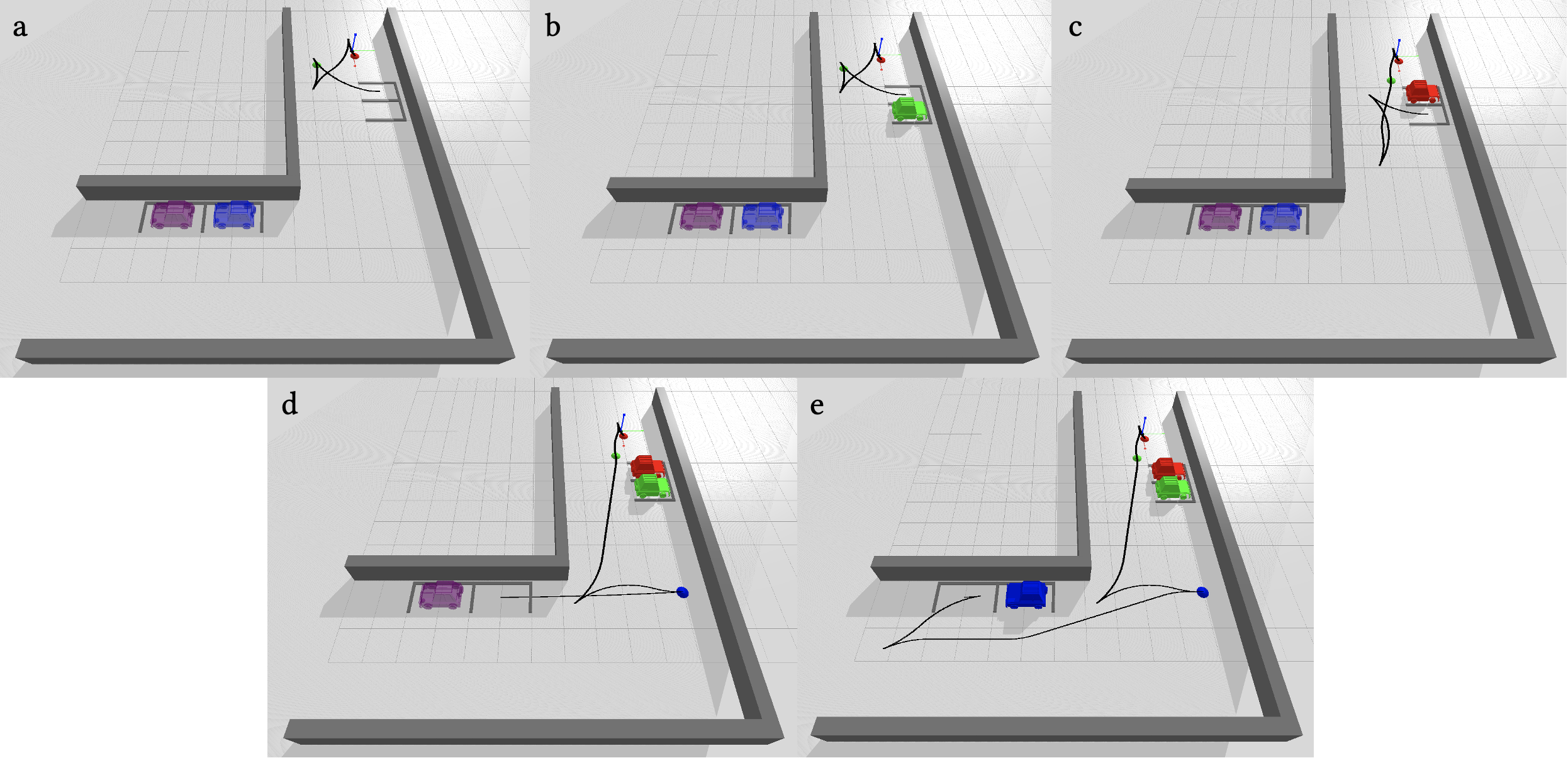}
    \caption{Result on parking environment with a Reeds-Shepp car.\label{fig:parking}}
\end{figure}

\section{Camera-based State Sampler\label{sec:camera_sampler}}

In \oldmethod, the random graph is constructed through uniform sampling. 
However, uniform sampling depends on chance whether a partially observable object is seen. 
Since the potential locations of all partially observable objects are known to the planner, we can take advantage of this, by developing a \emph{camera-based state sampler}.
This sampler creates configurations in which the robot's camera is directly looking towards a partially observable object that exists in the current world state. 

The camera-based state sampler is shown in Alg.~\ref{alg:camera}.
First, the algorithm calls the function \textsc{GetRandomObject} (Line 1), which returns a randomly selected object as the target object, i.e. the object that should be seen by the camera. Then the position of the target object is queried (Line 2), and a random workspace point is sampled (Line 3), at which we like our camera to be positioned. 

\begin{center}
\begin{algorithm}
\caption{Camera-based State Sampler\label{alg:camera}}
\RestyleAlgo{ruled}
\KwData{World $W$}
\KwResult{State $q$}
    $object$ $\gets$ GetRandomObject($W$)\;
    $T_{object} \gets$ GetObjectCenter(object)\;
    $T_{camera} \gets$ GetRandomWorkspacePoint()\;
    $c_z \gets T_{object} - T_{camera}$\;
    Normalize($c_z$)\;
    $e_z \gets$ [0, 0, 1]\;
    $c_x \gets$ CrossProduct($e_z$, $c_z$)\;
    $c_y \gets$ CrossProduct($v$, $c_x$)\;
    $R_{camera} \gets$ RotationMatrix([$c_x$, $c_y$, $c_z$])\;
    $q \gets$ CalculateInverseKinematics(\{$T_{camera}, R_{camera}$\})\;
    $\Return\ q$\;
\end{algorithm}
\end{center}

Given this information, we compute a camera frame. We first compute a normalized direction pointing towards the selected objects from the workspace point (Line 4, 5). 
Afterwards, a frame represented by a rotation matrix is computed, which points its $z$-axis towards the object, while keeping the camera horizontal (Line 6-9). 
This is done by first computing the $x$-axis of the frame as the cross-product of the direction $c_z$ and the unit vector that points in the positive $z$-direction (Line 6, 7). 
The $y$-axis of the frame consists of the cross-product of the direction and the $x$-axis of the frame (Line 8). This frame is represented as a rotation matrix located at the random workspace point (Line 9).

Finally, this camera frame is used in an inverse kinematics (IK) solver (Line 10). 
Based on the camera frame, the IK solver computes a joint configuration that leads to the camera pointing towards the target object. This configuration is eventually returned (Line 11).

This camera-based sampler can sample states from which the camera can directly observe partially observable objects. This leads to an increase in states in the random graph that are associated with observations. By combining this sampler with a uniform sampler, we can guarantee asymptotic optimality of PTO.

The camera-based sampler also implements a method that replaces the uniform sampling of the base position with values that can be passed to the method as arguments. This can be used in the random graph creation described in section \ref{sec:create_random}. Before sampling randomly, new states can be sampled at the corresponding start position of the base while the robot observes the existing objects in this configuration.
\begin{figure*}
    \centering
    \includegraphics[width=\linewidth]{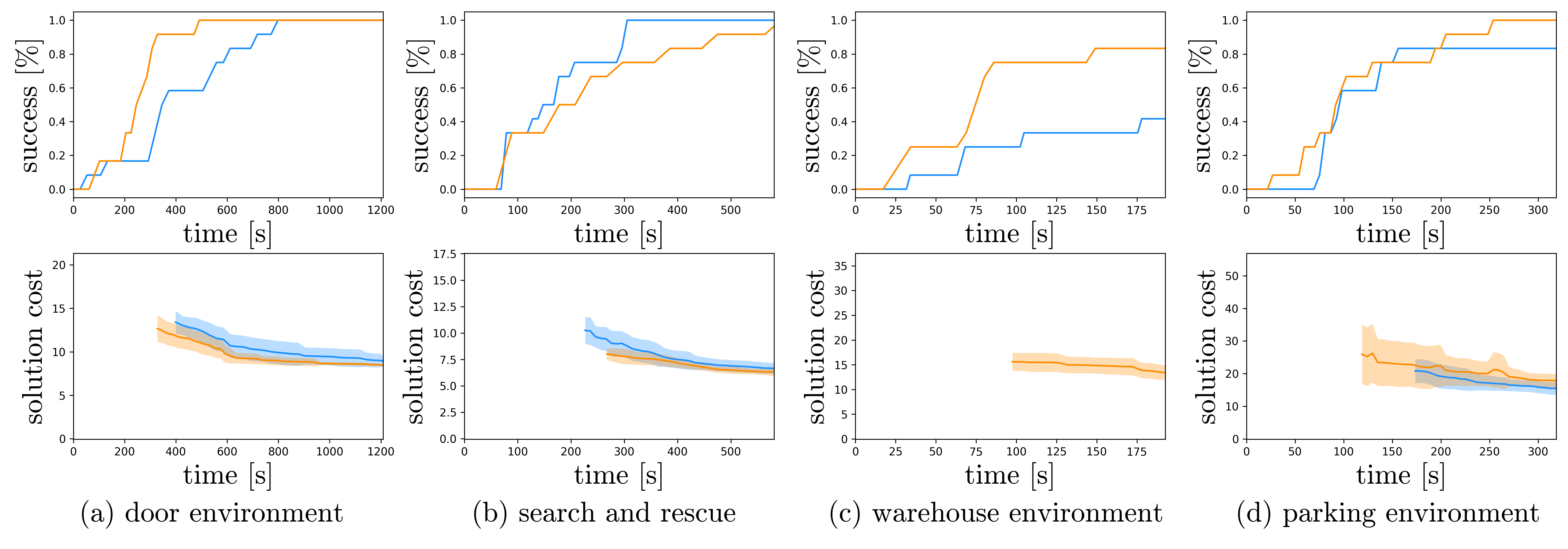}
    \caption{We compare \method with a uniform random state sampler~\sqbox{blue} against \method with the camera-based state sampler~\sqbox{orange}.\label{fig:benchmark}}
    \vspace*{-0.2cm}
\end{figure*}

\section{Experiments}

We evaluate our implementation of \method with the default sampler against PTO with the camera-based state sampler on four scenarios. 
Each run is repeated $10$ times and we report on average success rate and cost. 
The belief for all experiments is initialized with an uniformly distributed initial belief.
The experiments are performed on a laptop with 16 GB of RAM using Ubuntu 20.04. PyBullet is used in version 3.2.5. The OMPL library is extended using version 1.5. The code is open source and can be accessed via github\footnote{\url{https://github.com/janisfreund/path-tree-optimization}}. For each scenario, a different timeout has been used, which reflects realistic requirements, and makes the differences in solutions better visible. We showcase solution costs only after $50$\% of scenarios have been solved.

\subsection{Door Environment\label{sec:door}}

The first environment represents a living area with three doors (Fig.~\ref{fig:door_environment}). All doors are partially observable, i.e. the robot needs to observe if they are open or closed.

A resulting path tree is shown in Fig.~\ref{fig:door_environment}. 
Following the trajectory, the robot first observes the blue door.
If the door is open, the robot takes a path through the door and reaches the goal. 
If the blue door is closed, the robot next observes the red door. 
If the door is open, the robot directly moves to the goal. 
If the red door is closed, the robot instead moves until it can observe the green door. 
If it is open, the robot takes the path going through the green door. 
Otherwise, the robot takes the path avoiding all doors.

Figure \ref{fig:benchmark} shows the performance of the \method planner using the default sampler versus using the camera-based sampler. For the camera-based state sampler, the robot reaches a $100$\% success rate after around $500$ seconds, which is a $60$\% improvement to the around $800$ seconds needed for the default sampler. Both solution costs improve over time with a slight advantage for the camera-based state sampler. 

\subsection{Search and Rescue Environment\label{sec:sar}}

The third experiment is a search and rescue scenario. 
It consists of planning a path for a Franka robot arm \cite{FrankaEmikaPanda}, mounted on a mobile base. 
A camera is attached to the end-effector of the robot.
Together, the robot has ten dimensions, three corresponding to the base and seven corresponding to the arm. In the scenario, a dog must be rescued from an office. Potential locations of the dog are known in advance, but the actual position of the dog is unknown at planning time. 

A path tree, planned by the \method planner, is visualized in Figure \ref{fig:sar_environment}. Since the dog can only be at one location at once, there are exactly four different world states. First, the robot observes the potential locations of the robot in the following order: green, blue, purple, and red. If the observed location reveals the presence of the dog, the robot moves directly towards the site. Otherwise, it continues the path tree to the next observation point.

The results are shown in Fig.~\ref{fig:benchmark}.
Both state samplers find a solution after approximately $600$ seconds, but the camera-based sampler finds all solutions after $300$ seconds. 
The camera-based sampler converges relatively quickly with the standard deviation approaching zero, while the costs using the default sampler are higher and associated with a greater standard deviation. 
This can be explained by the $10$-dimensional configuration space, making it less likely to sample a state with an informative camera position.

\subsection{Warehouse Environment\label{sec:warehouse}}

The second experiment is an item retrieval task in a warehouse as shown in Fig.~\ref{fig:warehouse_environment}.
A path is planned for the Franka robot arm which is mounted on a mobile base. 
The location of the item is not known at planning time, but the planner knows about two potential positions. 
Both item locations are occluded by a wall, so the robot is not able to directly observe them.

A path tree planned by \method is shown in Fig.~\ref{fig:warehouse_environment}. 
There are only two possible states of the world: Either the object is at the red or the blue location. 
In the planned path tree, the robot directly moves towards the window. 
From this state, it is able to observe the blue item. 
If it exists, the robot moves through the left door and picks up the box. 
If the blue object does not exists, the robot moves through the right door.

The benchmark results are shown in Fig.~\ref{fig:benchmark}. In the allocated time frame, the camera-based state sampler solves $80$\% of the cases, while the default sampler solves $40$\%. 
Only the solution cost for the camera-based state sampler is shown (after $50$\% cases were solved). It can be seen that both the mean cost and the deviation decrease over time, indicating that the the planner converges to a low-cost solution.
This scenario demonstrates the benefits of using the camera-based sampler, since finding the window location is beneficial to allow the robot to quickly make the correct decision.

\subsection{Parking Environment\label{sec:parking}}

In the final environment, we evaluate \method on a parking scenario using the Reeds-Shepp car's state space~\cite{reeds} as shown in Fig.~\ref{fig:parking}. The task is to park in one of four parking lots. The parking lots can either be free or occupied.

Fig.~\ref{fig:parking} also shows a planned path tree. For better visibility, the branches of the path tree are shown in different images. 
First, the car observes the parking lot associated with the red car. 
If it is empty, the car drives towards this parking space, as seen in the images (a) and (b). 
If the red car exists, the car observes the parking lot of the green car next. 
If the green car does not exists, it directly parks in that lot as can be seen in image (c). 
If it exists, the car observes the spot of the blue car. 
It parks there if it cannot detect the blue car, as seen in image (d). Otherwise, it drives to the last remaining spot.

The benchmark results are shown in Fig.~\ref{fig:benchmark}. 
The samplers are comparable, although the default state sampler only find $80$\% of the solutions after $300$s. It is noticeable that the standard deviation of the cost is relatively high for both samplers, but converges over time to a similar value.
\section{Conclusion\label{sec:discussion}}

We presented \method, an improved planner based on prior work by Phiquepal et al.~\cite{pathtree}.
\method has been shown to solve problems with multiple partially observable objects, and problems with multiple partially observable goal regions. 
We implemented \method in the open motion planning library (OMPL), added support for simulated cameras, and for non-euclidean state spaces. An additional innovation is the novel camera-based state sampler, which biases sampling towards configurations at which important observations can be made. 
In our evaluations, we showed \method to converge to near optimal solutions.

While we believe this to be a significant improvement, there are two remaining limitations. 
First, partially observable objects can only take exactly two states. 
However, objects might be associated with more than two states, like reconfigurable tools, or different robot attachments. \method could accommodate this by adding an additional detection method of object states from the camera image.
Second, the \method method uses internally three sequentially executed phases. 
However, this can increase runtime because belief space sampling is decoupled.
This could be remedied by developing a unified version to directly sample in belief space.

Despite limitations, \method was shown to be a well-suited planner for motion planning in discrete partially observable environments.
\method has strong guarantees like asymptotic optimality, and is implemented in OMPL, such that the robotics community can benefit from and build upon this work.

\balance
\bibliographystyle{plain}
\bibliography{bib/references}

\end{document}